\documentclass{article} 
\usepackage{iclr2026_conference,times}
\usepackage{subcaption}
\usepackage{caption}

\usepackage{amsmath,amsfonts,bm}

\def\eqref#1{equation~\ref{#1}}

\def\1{\bm{1}}

\DeclareMathAlphabet{\mathsfit}{\encodingdefault}{\sfdefault}{m}{sl}
\SetMathAlphabet{\mathsfit}{bold}{\encodingdefault}{\sfdefault}{bx}{n}

\usepackage{graphicx}
\usepackage{hyperref}
\usepackage{url}
\usepackage{enumitem}
\usepackage{graphicx}
\usepackage{booktabs} 
\title{How to model Human Actions distribution with Event Sequence Data}

\author{Egor Surkov, Dmitry Osin, Evgeny Burnaev, Egor Shvetsov \\
Applied AI\\
Moscow, Russia \\
\texttt{egorsurkov.ds@gmail.com, dima.tina2013@gmail.com,}\\
\texttt{e.burnaev@applied-ai.ru, geksut@gmail.com}
}

\iclrfinalcopy 
\begin{document}

\maketitle

\begin{abstract}
This paper studies forecasting of the future distribution of events in human action sequences, a task essential in domains like retail, finance, healthcare, and recommendation systems where the precise temporal order is often less critical than the set of outcomes. We challenge the dominant autoregressive paradigm and investigate whether explicitly modeling the future distribution or order-invariant multi-token approaches outperform order-preserving methods. We analyze local order invariance and introduce a KL-based metric to quantify temporal drift. We find that a simple explicit distribution forecasting objective consistently surpasses complex implicit baselines. We further demonstrate that mode collapse of predicted categories is primarily driven by distributional imbalance. This work provides a principled framework for selecting modeling strategies and offers practical guidance for building more accurate and robust forecasting systems.
\end{abstract}

\section{Introduction}

In many real-world prediction tasks, the precise temporal ordering of events is irrelevant. Instead, predicting the distribution of outcomes, where only the presence or absence of specific elements matters, is sufficient and often more practical.

For instance, in retail operations, probabilistic demand forecasting enables optimal inventory management and supply chain planning by modeling the full range of possible product demands without requiring sequence order~\citep{nassibi2023demand, larson2001designing}. Similarly, in healthcare, clinical diagnosis systems treat disease categories as unordered sets within a single hospital admission. The presence of certain  conditions is clinically more significant than the exact order in which they were diagnosed~\citep{johnson2016mimic, mullenbach2018explainable}. Recommendation systems further exemplify this principle known as \textit{basket prediction}~\citep{rendle2020recommender}. Finally, many multi-label problems can naturally be framed as distribution forecasting tasks.

\textbf{The central focus of this paper}~\textit{is to model the future distribution of human actions over a fixed future horizon.} In this work we consider \textit{Event Sequences} (EvS) \citep{osin2025ebes, udovichenko2024seqnas} - temporal records of human actions which underpin a wide range of decision-making systems across domains including healthcare~\citep{johnson2016mimic}, financial transactions~\citep{udovichenko2024seqnas, mollaev2024multimodal, yang2019fraudmemory}, e-commerce~\citep{li2021novel}, recommender systems~\citep{shevchenko2024variability,klenitskiy2024does, zhelnin2025faster}, and human action recognition~\citep{surkov2024human}. Despite its practical importance and deceptively simple formulation, distribution forecasting for \textbf{EvS} remains significantly understudied.

Inspired by advances in Natural Language Processing (NLP), contemporary approaches to modeling sequential behavior often default to autoregressive generation predicting the next token conditioned on an exact prefix ordering~\citep{karpukhin2024hotpp, klenitskiy2024does}. 
While Next Token Prediction (NTP) has long dominated sequential modeling, Multi-Token Prediction (MTP) has recently gained traction due to its demonstrated improvements in model quality and generalization, particularly in tasks such as planning, code generation and EvS forecasting~\citep{nagarajan2025roll,bachmann2024pitfalls, yu2025autoregressive, karpukhin2024detpp}.


\textbf{This raises a practical question:} When should we model future event distributions \textbf{explicitly}, and when is it worth preserving temporal structure through implicit methods like NTP or Multi-Token Prediction (MTP)?

Two challenges complicate this choice. First, event sequences often exhibit \textit{local order invariance}: within short time windows, the precise ordering of actions (e.g., “buy bread” vs. “buy aspirin”) may be arbitrary or uninformative, especially in transactional domains \citep{klenitskiy2024does, osin2025ebes, udovichenko2024seqnas}. Second, as we demonstrate empirically in this work, NTP frequently suffers from \textit{mode collapse} on certain datasets.
We investigate three hypotheses for this collapse:  
\label{sec:hypo}

\begin{itemize}[leftmargin=*, itemsep=0pt, parsep=0pt]
    \item \textbf{H1 (Global Order Irrelevance)}: Full sequence order is uninformative. 
    \item \textbf{H2 (Local Order Irrelevance)}: Only coarse temporal structure matters; fine-grained order is uninformative. NTP, by overemphasizing local order, may fail to capture these higher-order distributional patterns.
    \label{sec:h2}

    \item \textbf{H3 (Distributional Imbalance)}: Highly skewed label distributions can independently cause mode collapse — especially under maximum likelihood training with NTP, which tends to overweight frequent tokens.

\end{itemize}


To resolve these questions, we introduce a \textit{diagnostic-driven framework} for distribution forecasting:
\begin{itemize}[leftmargin=*, itemsep=0pt, parsep=0pt]

\item \textbf{Quantifying Temporal Drift:} We introduce a \textbf{KL-based staticity index}, which quantifies distributional drift over time ,providing a dataset-level diagnostic for the relevance of global temporal structure.

\item \textbf{Local Order Invariance:} We conduct controlled experiments in which we randomly permute events within sliding windows of varying lengths during to training and measure each dataset’s sensitivity to local and global order disruption. Moreover, we provide a formal analysis demonstrating that \textbf{NTP} would fail with locally invariant data structure.

\item \textbf{Mode Collapse Analysis:} We quantify distribution characteristics for each dataset, such as the exponential decay factor. 
Although we did not observe any correlation between mode collapse and distribution skewness, our results indicate connection of skewness and sensetivity to invariance.


\item \textbf{Explicit vs. Implicit Objectives:} We systematically compare four training paradigms: (1) Next Token Prediction (NTP), (2) Multi-Token Prediction (MTP) with ordered output, (3) an order-invariant set prediction approach with post-hoc alignment to targets, and (4) a novel explicit distribution forecasting objective that directly models category probabilities without enforcing order. Our experiments show that order-invariant approaches, particularly the simple explicit method significantly outperform order preserving baselines across most domains. Surprisingly, even on textual data — where order is traditionally assumed critical the explicit approach remains superior.


\end{itemize}

We evaluate all methods across two models and seven public datasets spanning recommendation, retail, banking, and natural language domains.
By unifying these findings, our work provides actionable practitioner guidance and paves the way for future research in human behavior modeling.

\section{Related Work}
\paragraph{Architectures for Event Sequences.}
Modeling user actions sequentially by conditioning on past behavior has become an essential component of modern recommendation pipelines. These approaches effectively adapt ideas from natural language processing (NLP), particularly attention-based architectures~\citep{kang2018self, sun2019bert4rec, klenitskiy2024does, mezentsev2024scalable}. However, it remains unclear whether transformer-based architectures are indeed the most suitable for predicting future user actions. In \emph{EBES}~\cite{osin2025ebes} and in  \emph{SeqNAS}~\cite{udovichenko2024seqnas}, the authors demonstrate that RNN-based architectures outperform transformer-based models on \textbf{EvS} classification tasks. Delving deeper into this issue, \cite{karpukhin2025ht} investigate the limitations of transformers and proposes several modifications that enable them to surpass RNNs in classification performance. However, as the same work further reveals, these enhancements do not translate to improved performance in forecasting future tokens. In this work, we focus on RNN- and GPT-based architectures, as they remain the most applicable in this domain.

\paragraph{Multi-Token vs. Single-Token Prediction.}
Multi-Token Prediction (MTP) has recently gained traction due to its demonstrated improvements in model quality and generalization particularly in tasks such as planning, code generation~\citep{nagarajan2025roll,bachmann2024pitfalls,yu2025autoregressive}. However, a key challenge lies in the common assumption that predicted tokens are conditionally independent~\cite{gloeckle2024better}.

\emph{Teacherless Learning}~\cite{bachmann2024pitfalls} offers an intermediate approach between Next-Token Prediction (NTP) and MTP, conceptually opposing teacher forcing. Unlike MTP, Teacherless Learning is grounded in a rigorous mathematical framework. While it does not accelerate inference, it addresses fundamental limitations of traditional NTP. As \cite{nagarajan2025roll} note: “Teacherless training and diffusion models comparatively excel in producing diverse and original output.” 

Although earlier work focused primarily on text generation, \cite{karpukhin2024detpp} extended these ideas to \textbf{EvS}, demonstrating that multi-token generation and diffusion-based approaches indeed outperform the single-token paradigm. In this work, we investigate NTP, a multi-token strategy similar to that proposed in \cite{karpukhin2024detpp} and propose a new explicit approach for distribution forecasting. 

\paragraph{Order Importance in \textbf{EvS}.}

It has been established that permuting sequences in \textbf{EvS} datasets does not degrade performance on classification tasks~\citep{osin2025ebes,moskvoretskii2024self}, an observation  which significantly challenges the assumed sequential nature of this data type. \cite{klenitskiy2024does} investigates whether datasets from the domain of sequential recommender systems genuinely exhibit sequential structure. Specifically, the authors evaluate whether permuting sequences leads to performance degradation in next-token prediction tasks, and find that the extent of degradation varies by dataset, some datasets are more “sequential” than others. In this work, we extend this investigation beyond recommender systems and analyze local permutation invariance, as discussed in \textbf{H2}~(Section~\ref{sec:h2}).

\section{Datasets}
To evaluate the proposed methods and hypotheses, we conduct experiments on a diverse collection of real-world sequential datasets spanning multiple domains—including financial transactions, e-commerce, retail, music streaming, and literary text. A summary of key statistics is provided in Table~\ref{tab:datasets}; full descriptions, including preprocessing steps are available in Appendix~\ref{app:datasets}.

\begin{table}[htbp]
\centering
\caption{Dataset statistics and characteristics.}
\label{tab:datasets}
\resizebox{\textwidth}{!}{%
\renewcommand{\arraystretch}{1.3}
\begin{tabular}{@{}lllllll@{}}
\toprule
\hline
\textbf{Dataset} & \textbf{ID} & \textbf{Domain} & \textbf{Sequences} & \textbf{Mean len} & \textbf{Target Field} & \textbf{Classes}\\
\hline
Multimodal Banking Dataset~\citeyear{mollaev2024multimodal} 
    & MBD
    & Transactions 
    & 1.5M
    & 313
    & Event type
    & 55 \\

AgeGroup Transactions
    & AGE
    & Transactions 
    & 30K
    & 888
    & Small group
    & 203 \\

X5 RetailHero 
    & Retail
    & Retail 
    & 40K
    & 112
    & Level 2
    & 43 \\
Alphabattle-2.0 
    & AB
    & Transactions
    & 1M
    & 213
    & MCC category
    & 28 \\
Complete Works of Shakespeare
    & ShS
    & Text 
    & 5K
    & 106
    & Character
    & 65 \\

Megamarket~(\citeyear{shevchenko2024variability})
    & MM
    & E-commerce
    & 2.73M
    & 653
    & Category ID
    & 9.8K \\

Zvuk~(\citeyear{shevchenko2024variability}) 
    & Zvuk
    & Music Streaming 
    & 380K
    & 1020
    & Artist ID
    & 210K \\

Taobao User Behavior
    & Taobao
    & E-commerce
    & 10K
    & 535
    & Item category
    & 8K \\
\hline
\end{tabular}%
}
\end{table}

\section{Dataset diagnostic}

\subsection{Temporal Order and Mode Collapse in Event Sequence Modeling}
\label{sec:mode_collapse_theory}
In time series and natural language modeling, precise temporal ordering is crucial. However, in domains like system logs or bank transactions, the \emph{exact micro-temporal order} of events within short windows may be ambiguous or irrelevant—e.g., two unrelated log entries milliseconds apart could plausibly appear in either order without changing system semantics. We illustrate this effect in Appendix~\ref{fig:order_example}. This motivates a formal distinction between two types of temporal structure:
\begin{itemize}[left=2pt]
    \item \textbf{Local invariance}: Within a narrow window $W_t = (y_t, \dots, y_{t+H})$, event order is semantically irrelevant—permutations of the same multiset are equally plausible.

    \item \textbf{Global structure}: Across broader time intervals, dependencies between consecutive windows remain meaningful; e.g., $p(W_2 \mid W_1)$ for $W_1 = (y_0, \dots, y_{t-1})$ and $W_2 = (y_t, \dots, y_{t+H})$ captures genuine temporal progression.

\end{itemize}

Conventional autoregressive (AR) models are trained to predict the next token $y_t$ given its full history $(y_0, \dots, y_{t-1})$. To accommodate local invariance, one might relax this strict left-to-right dependency by defining a \emph{prediction horizon} $\{y_t, \dots, y_{t+H}\}$ and training the model to predict \emph{any} event within this window. Under the assumption of uniform uncertainty over the horizon, the training objective becomes:

\begin{align}
\mathbb{E}_{k \sim \text{Uniform}[0,H]} 
\big[ \log p(x = y_{t+k} \mid y_0, \dots, y_{t-1}) \big] 
= \frac{1}{H+1} \sum_{m=0}^{H} \log p(x = y_{t+m} \mid y_0, \dots, y_{t-1}).
\end{align}


Critically, standard AR architectures use a \emph{single output distribution} $q_t(\cdot)$ at time $t$ to score all tokens in the horizon. Under local permutation invariance, the optimal $q_t$ that maximizes the above objective is the empirical distribution over the multiset $\{y_t, \dots, y_{t+H}\}$. Consequently, the model learns a \emph{static predictive distribution} over the entire window: $q_t \approx q_{t+1} \approx \cdots \approx q_{t+H}$.
This static distribution becomes problematic at inference time. When generating sequences using deterministic decoding (e.g., argmax or low-temperature sampling), the model outputs:
\[
\hat{y}_{t+k} = \arg\max_x q_{t+k}(x) \approx \arg\max_x q_t(x), \quad \forall k \in [0, H].
\]
Since $q_t$ is dominated by the most frequent event in the window, the model repeatedly predicts the \emph{empirical mode} of $\mathcal{W}_t$, suppressing rarer—but valid—events. We term this phenomenon \emph{temporal mode collapse}.



We propose that explicitly modeling the distribution of events across entire windows, rather than enforcing pointwise predictions, offers a principled resolution. This allows models to better capture the stochastic nature of real-world event sequences while avoiding degenerate solutions.

\subsection{Staticity index}

Before fitting neural models, we quantify how each sequence’s event
distribution changes over time.  Several datasets contain sequences with nearly static behaviour; to verify this, we plot the \emph{Shape} score drift for each dataset. 

\subsubsection{Per-feature dissimilarity score}\label{subsec:shape}

\textbf{Procedure.}
For each sequence, we fix a window length \(W\) and stride \(s\), then slide
the window across the timeline.  At every position \(i\), we extract the
feature distribution \(P_i\) within the current window and compare it with
the baseline distribution \(P_0\) computed from the first window. To compare them we suggest to leverage the following score:

Let \(P_0\) and \(P_i\) denote the empirical distributions in the reference
and the \(i\)-th window, respectively.

\textbf{Discrete features.}
For categorical attributes defined on \(\mathcal{A}\) we employ the
\emph{total variation (TV) distance}, 
$
\mathrm{TV}(P_0,P_i)
  \;=\;
  \tfrac12
  \sum_{a\in\mathcal{A}}
  \bigl|P_0(a)-P_i(a)\bigr|.
$
Because lower $\mathrm{TV}$ indicates higher similarity, we report its complement $(1-\mathrm{TV})$, so that higher values consistently reflect better alignment.

\textbf{Continuous features.}
For numerical attributes we use the \emph{Kolmogorov–Smirnov} statistic.
Let \(F_0\) and \(F_i\) be the empirical CDFs corresponding to \(P_0\) and
\(P_i\).  The KS divergence is
 $
 \mathrm{KS}(P_0,P_i)
  \;=\;
  \sup_{x\in\mathbb{R}}
  \bigl|F_0(x)-F_i(x)\bigr|.
  $
Analogously, we report the similarity score \(1-\mathrm{KS}\).

\paragraph{Shape score.}
For window \(i\) we propose to compute each feature’s distance using the appropriate formula above and then average across all features: $
  \mathrm{Shape}(P_0,P_i)
  \;=\;
  \frac{1}{M}
  \sum_{j=1}^{M}
  d_j\!\bigl(P_0^{(j)},P_i^{(j)}\bigr),
$
where \(M\) is the number of features and \(d_j\) is
\(\mathrm{TV}\) when the $j$th feature is categorical, and \(\mathrm{KS}\) otherwise.  Plotting \(i\mapsto\mathrm{shape}(P_0,P_i)\) yields the drift
curves used throughout this paper.

With these definitions we shift the window across the entire sequence and plot trajectory \(i \mapsto \mathrm{shape}(P_0,P_i)\), obtaining time-resolved drift curve that summarises how the distribution evolves over the time.

\subsubsection{Staticity in Datasets}

Across banking datasets (MBD, Retail, Age, AlphaBattle) the majority of user sequences form static clusters with negligible temporal drift (Figure~\ref{fig:staticity-mbd}, Appendix~\ref{appendix:staticity}). In contrast, RecSys data such as ZVUK exhibit more diverse and volatile trajectories (Figure~\ref{fig:staticity-zvuk}, Appendix~\ref{appendix:staticity}), while the Shakespeare dataset, despite being textual, resembles banking data with largely flat drift patterns~(Figure~\ref{fig:staticity-shakespeare}, Appendix~\ref{appendix:staticity}). Detailed analyses for individual datasets are provided in the Appendix~\ref{appendix:staticity}.

\paragraph{Motivation.}
These observations motivate a prevent-level \emph{staticity index} that can be computed \emph{before} model training to guide the choice of modeling strategy.  
Unlike the single–anchor variant (first window vs. all others), we adopt a more robust, multi–anchor formulation.

\textbf{Staticity index.}
Fix a window length \(W\) and stride \(s\). 
For each sequence \(u\) with per–window distributions \(\{P_i^{(u)}\}_{i=1}^{I_u}\), 
choose anchors \(\mathcal{R}_u\) (uniformly at random, \(R=3\)). 
The per–sequence score is the average shape–similarity
\[
S^{(u)} = \frac{1}{R I_u} \sum_{r\in\mathcal{R}_u} \sum_{i=1}^{I_u}
          \mathrm{Shape}\ \!\bigl(P_r^{(u)}, P_i^{(u)}\bigr),
\]
and the dataset–level index is $\mathrm{Staticity} = \tfrac{1}{N}\sum_{u=1}^N S^{(u)}.$

\begin{figure}[t]
\centering
\begin{minipage}{0.4\textwidth}
  \centering
  \includegraphics[width=\linewidth]{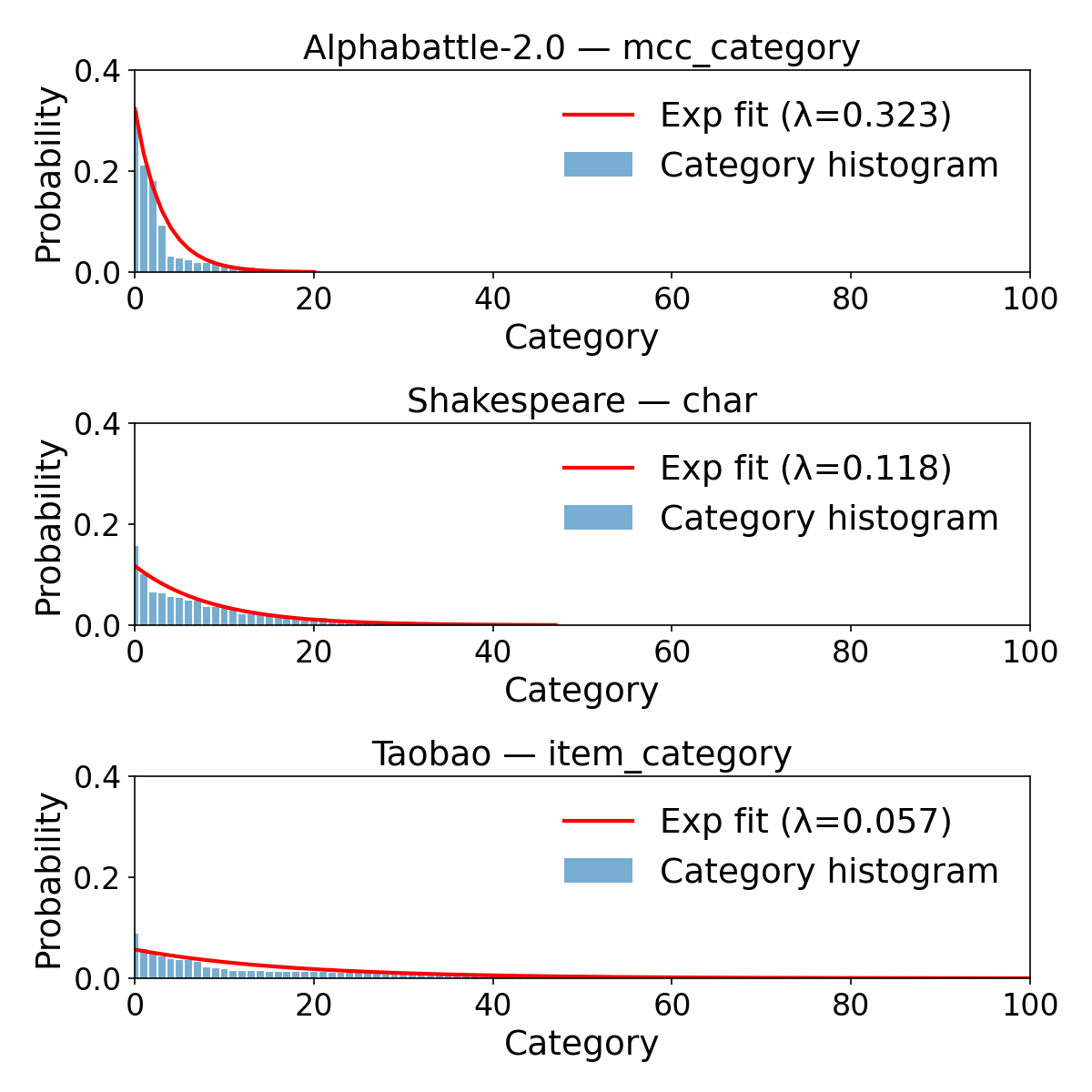}
  \caption{Distribution of categories in datasets. We present normalized number of categories.}
\end{minipage}\hfill
\begin{minipage}{0.55\textwidth}
  \centering
    \captionof{table}{Dataset statistics: exponential decay parameter ($\lambda$), Staticity index ($S$), total number of distinct categories (TCD), and perplexity (PPL) increase after full shuffle.}

\begin{tabular}{lcccc}
    \hline
    \multicolumn{1}{c}{\bf Dataset} & \multicolumn{1}{c}{\boldmath$\lambda$} & \multicolumn{1}{c}{\bf TCD} & \multicolumn{1}{c}{\bf $\mathbf{S}$} & \multicolumn{1}{c}{\bf PPL} \\
    \hline
    \multicolumn{5}{c}{\textit{Banking domain}} \\
    \hline
    MBD     & 0.415 & 55   & 0.842 & 1.02$\times$ \\
    AB      & 0.305 & 28   & 0.725 & 1.08$\times$ \\
    Age     & 0.245 & 203  & 0.772 & 1.24$\times$ \\
    Retail  & 0.185 & 43   & 0.782 & 1.27$\times$ \\
    \hline
    \multicolumn{5}{c}{\textit{Text}} \\
    \hline
    ShS     & 0.118 & 64   & 0.803 & 5.09$\times$ \\
    \hline
    \multicolumn{5}{c}{\textit{Recommender Systems}} \\
    \hline
    Taobao  & 0.016 & 1.9K & 0.650 & 13.00$\times$ \\
    MM      & 0.005 & 9.8K & 0.406 & 10.87$\times$ \\
    Zvuk    & 0.003 & 210K & 0.363 & 5.05$\times$ \\
    \hline
\end{tabular}


  \label{tab:dataset_statistic}
  
\end{minipage}
\end{figure}

Thus, the staticity index quantifies the temporal stability of a sequence’s multi-feature distribution: higher values (close to 1) reflect stronger staticity (quasi-stationarity), whereas values near zero indicate pronounced drift. Importantly, the conclusions derived from the computed staticity index (Table~\ref{tab:dataset_statistic}) align with those previously inferred from the qualitative analysis of the plots.

\subsection{Local Permutation of Events}
To assess the role of temporal order, we permute events within a symmetric window of some size centered at each position. We evaluate different windows, where $w=\{0,1,4,16,-1\}$ denotes a number of permuted neighbors, $w=-1$ corresponds to permutation of the full sequence. This design reveals whether models rely on local ordering or global sequence structure.

\section{Distribution Forecasting Methods}

We study the task of forecasting a distribution of a sequence over some horizon $N$ given its history. To this end, we consider several training objectives — autoregressive, target-based, matched, and our order-invariant formulation.  For all experiments $N$ is fixed as $32$.

\subsection{Autoregressive loss}

Let $x_{1:T}$ be a sequence with $x_t\in\{1,\dots,K\}$. The model parameterises conditional next–event probabilities
$p_\theta(x_{t+1}\mid x_{1:t})$ given the preceding context $x_{1:t}$.
The sequence log–likelihood factorises as:
  $\log p_\theta(x_{1:T})
  \;=\;
  \sum_{t=1}^{T}
  \log p_\theta\!\bigl(x_{t+1}\mid x_{1:t}\bigr).$
\subsection{Target loss}

In this setting the model predicts an entire block of \(L\) future events
\emph{in a single forward pass}, using a fixed prefix
\(x_{1:T}\) as context; no teacher forcing is applied inside the
horizon.  
Let \(\hat p_{T+1},\dots,\hat p_{T+L}\) be the categorical
distributions produced for positions \(T\!+\!1\) through \(T\!+\!L\).
The target loss is the sum of negative log-likelihoods for that block: $\mathcal{L}_{\text{target}}^{(L)}
= \sum_{i=T+1}^{T+L}
    -\log \hat p_{\,i}\!\bigl(x_{i}\,\bigl|\,x_{1:T}\bigr)
$

Unlike the autoregressive objective, every term is conditioned on
\emph{the same} prefix \(x_{1:T}\); the model \textbf{GRU-Target} therefore learns to produce an
entire horizon coherently without receiving the ground-truth events
\(x_{T+1:T+L-1}\) as intermediate inputs.

\subsection{Matched loss}

When the temporal order of future events is weakly informative, forcing the
model to predict both the \emph{events} and their \emph{exact positions} needlessly penalises near-correct outputs. The \textbf{GRU-Matched} model adapts the matching idea of \cite{karpukhin2024detpp}, aligning each target event with the nearest prediction within a tolerance window of size~\(m\), treated as a hyperparameter.

Let a fixed prefix \(x_{1:T}\) condition a one-shot block prediction
\(\hat p_{T+1:T+L}\); let \(x_{T+1:T+L}\) be the corresponding ground truth.  
With a permutation \(\sigma\) constrained by \(|\sigma(i)-i|\le m\),
the matched loss is $
\mathcal{L}_{\text{match}}^{(m)}
=
\min_{\substack{\sigma\in \mathcal{A}\\|\sigma(i)-i|\le m}}
\;
\sum_{i=T+1}^{T+L}
  -\log \hat p_{\sigma(i)}\!\bigl(x_i \mid x_{1:T}\bigr).
$

At \(m=0\) it reduces to plain block cross-entropy; as \(m\) grows, the objective becomes progressively order-invariant.  The minimisation is solved with the Hungarian algorithm on the cost matrix
\(\ell_{ij}=-\log\hat p_{j}(x_i \mid x_{1:T})\).

\subsection{Order-Invariant Distribution Parameterization}

When the order of future events is not informative, it is sufficient to model only the \emph{event type distribution} rather than their precise temporal arrangement.  
We therefore introduce the \textbf{GRU-Dist} model, which represents each sequence as a \emph{bag of events} and is trained to match the empirical distribution.

Let $H_t=\{x_1,\dots,x_t\}$ be the multiset of events observed so far. 
A neural encoder $f_\theta$ maps $H_t$ to logits, which are converted to probabilities
$ \boldsymbol\pi_t = \mathrm{softmax}\bigl(f_\theta(H_t)\bigr)\
  ;\in\;\Delta^{K-1}
$, where $\Delta^{K-1}$ is the probability simplex in $\mathbb{R}^K$.
For a sequence of length $L$ we form its empirical distribution
$
  \hat p_k=\tfrac{1}{L}\sum_{t=1}^L \mathbf{1}\{x_t=k\},
$
and minimize $\ell(\theta)=\mathrm{D_{KL}}\bigl(\hat p \,\|\, \boldsymbol\pi(\theta)\bigr).$

Unlike autoregressive objectives that require $L\times K$ logits per sequence, our order-invariant head outputs only a single $K$-dimensional vector. 
This reduces both computational and memory costs by a factor of $L$, while remaining well suited for datasets where event order carries little information.

\subsection{Multi-Token Prediction via Sampling}

Autoregressive decoding with greedy argmax often collapses to the modal category. 
A simple remedy is to \emph{sample} from the predictive categorical distribution instead of always taking the maximum, which reduces mode collapse and improves order-invariant metrics. 
For autoregressive and block-prediction models this sampling is straighforward, as logits at each step define the distribution, in our order-invariant method the distribution itself is parameterized directly, making sampling the natural decoding mechanism. We did not analyze more sophisticated sampling approaches such as beam search and our preliminary experiments with temperature sampling did not provide stable improveent across datasets, so we do not use them. \textbf{Sampling in the order-invariant model:}  Given a predicted categorical distribution $\pi=(\pi_1,\dots,\pi_K)$ and a target length \(L\), we compute expected counts  
$n_k = L \cdot \pi_k, \sum_{k=1}^K n_k = L$.
The category counts are computed using Hamilton’s method from apportionment theory~\cite{balinski2010fair}, which distributes $L$ discrete slots among categories in proportion to their predicted probabilities $\pi_k$ and guarantees that the total count equals $L$. 

\section{Evaluation}
For each configuration $Dataset \times Method \times LocalShuffle $ we  perform an extensive hyperparameter optimization of $100$ trails, technical details  are given in Appendix~\ref{app:hpo}.
\subsection{Baselines}
\textbf{We consider four simple baselines.} \textbf{(1) Ground Truth} uses the original sequences as a sanity check and reference point for metrics such as Cardinality. \textbf{Repeat} extends a sequence by copying its most recent observations into the forecast horizon of the lenght $N$. \textbf{Mode} outputs the users most frequent category for all $N$, illustrating the tendency of autoregressive models to collapse into trivial mode repetition—a behavior that may be overestimated by order-dependent metrics (e.g., Accuracy, Levenshtein distance).
Finally, \textbf{HistSampler} generates sequences by sampling from the empirical histogram of past users sequence, thereby preserving marginal category frequencies while discarding temporal dependencies.

\subsection{Metrics}

Many classical sequence metrics (e.g., Accuracy, Levenshtein distance, F1-score) are defined with respect to a fixed token order and therefore penalize any permutation of events, even when such reordering is irrelevant for the problem at hand. To overcome this limitation, we introduce an order-invariant \textit{Matched\mbox{-}F1} score, which treats sequences as \emph{bags of events}.

To avoid order dependence we redefine true-positive, false-positive and false-negative terms. Let $g_k$ and $\hat g_k$ denote the ground‑truth and predicted multiplicities of
class~$k$ in the window. We set

\[
(\text{TP}_k,\, \text{FP}_k,\, \text{FN}_k)
= \bigl(\min(g_k,\hat g_k),\;\max(0,\hat g_k-g_k),\;\max(0,g_k-\hat g_k)\bigr).
\]

Based on this definitions, we compute our \textit{Matched\mbox{-}F1} with \textbf{micro-} and \textbf{macro-}averaging, analogous to the conventional F1 formulation. Detailed definition of this metric placed in Appendix~\ref{appendix:f1}

To assess diversity, we use \textbf{Cardinality} (see Appendix~\ref{appendix:cardinality}), which measures the number of distinct categories generated by the model. Low values signal mode collapse, while values close to the ground-truth indicate faithful event variety. For completeness, we also report \textbf{Levenshtein distance}, an order-sensitive metric that, although less relevant to our setting, provides a complementary reference for order preserving methods.


\begin{table}[t]
\caption{Next $N$ tokens forecasting. \textit{Matched\mbox{-}F1 (micro)} for all datasets and methods including baselines. \textsuperscript{$\dagger$}~denotes sampled version of method.}
\label{tab:f1-micro-all-datasets}
\begin{center}
\begin{tabular}{lcccccccc}
\multicolumn{1}{c}{\bf Method} & \multicolumn{1}{c}{\bf MBD} & \multicolumn{1}{c}{\bf Age} & \multicolumn{1}{c}{\bf AB} & \multicolumn{1}{c}{\bf Retail} & \multicolumn{1}{c}{\bf ShS} & \multicolumn{1}{c}{\bf Taobao} & \multicolumn{1}{c}{\bf MM} & \multicolumn{1}{c}{\bf Zvuk} \\
\hline \\
GT          & 1.000 & 1.000 & 1.000 & 1.000 & 1.000 & 0.926 & 1.000 & 1.000 \\
Mode        & 0.520 & 0.331 & 0.380 & 0.219 & 0.158 & 0.117 & 0.156 & 0.113 \\
Repeat      & 0.830 & 0.680 & 0.700 & 0.661 & 0.587 & \textbf{0.257} & \textbf{0.318} & \textbf{0.274} \\
HistSampler & 0.804 & 0.632 & 0.680 & 0.640 & 0.533 & 0.197 & 0.244 & 0.226 \\

\hline
GRU & 0.528 & 0.477 & 0.375 & 0.207 & 0.596 & 0.222 & 0.250 & 0.148 \\
GRU\textsuperscript{$\dagger$} & 0.771 & 0.628 & 0.641 & 0.609 & 0.596 & 0.146 & 0.171 & 0.126 \\
\hline
GPT & 0.524 & 0.476 & 0.373 & 0.212 & 0.594 & 0.223 & 0.250 & 0,192 \\
GPT\textsuperscript{$\dagger$} & 0.776 & 0.627 & 0.629 & 0.611 & 0.603 & 0.151 & 0.188 & 0,174 \\
\hline
GRU-Target & 0.541 & 0.370 & 0.403 & 0.398 & 0.299 & 0.196 & 0.267 & 0.143 \\
GRU-Target\textsuperscript{$\dagger$} & 0.808 & 0.633 & 0.670 & 0.641 & 0.572 & 0.154 & 0.201 & 0.140 \\
\hline
GRU-Matched & 0.847 & 0.704 & 0.676 & 0.708 & 0.688 & 0.203 & 0.272 & 0.202 \\
GRU-Matched\textsuperscript{$\dagger$} & 0.827 & 0.653 & 0.647 & 0.667 & 0.634 & 0.155 & 0.203 & 0.134 \\
\hline
GRU-Dist & \textbf{0.856} & \textbf{0.725} & \textbf{0.736} & \textbf{0.719} & \textbf{0.705} & 0.178 & 0.247 & 0.239 \\
\end{tabular}
\end{center}
\end{table}

\section{Results}
\textbf{Dataset-level statistics.}
The staticity index serve as useful diagnostics for anticipating whether sequence order is relevant. Results are presented in Table~\ref{tab:dataset_statistic}. In banking datasets, a single modal category dominates—accounting for more than 50\% of all events—leading to high values of both $\lambda$ and the staticity index. This dominance is also associated with a pronounced performance drop under local permutations, suggesting limited reliance on sequential order.

\begin{figure}[h]
\begin{center}
\includegraphics[width=1\linewidth]{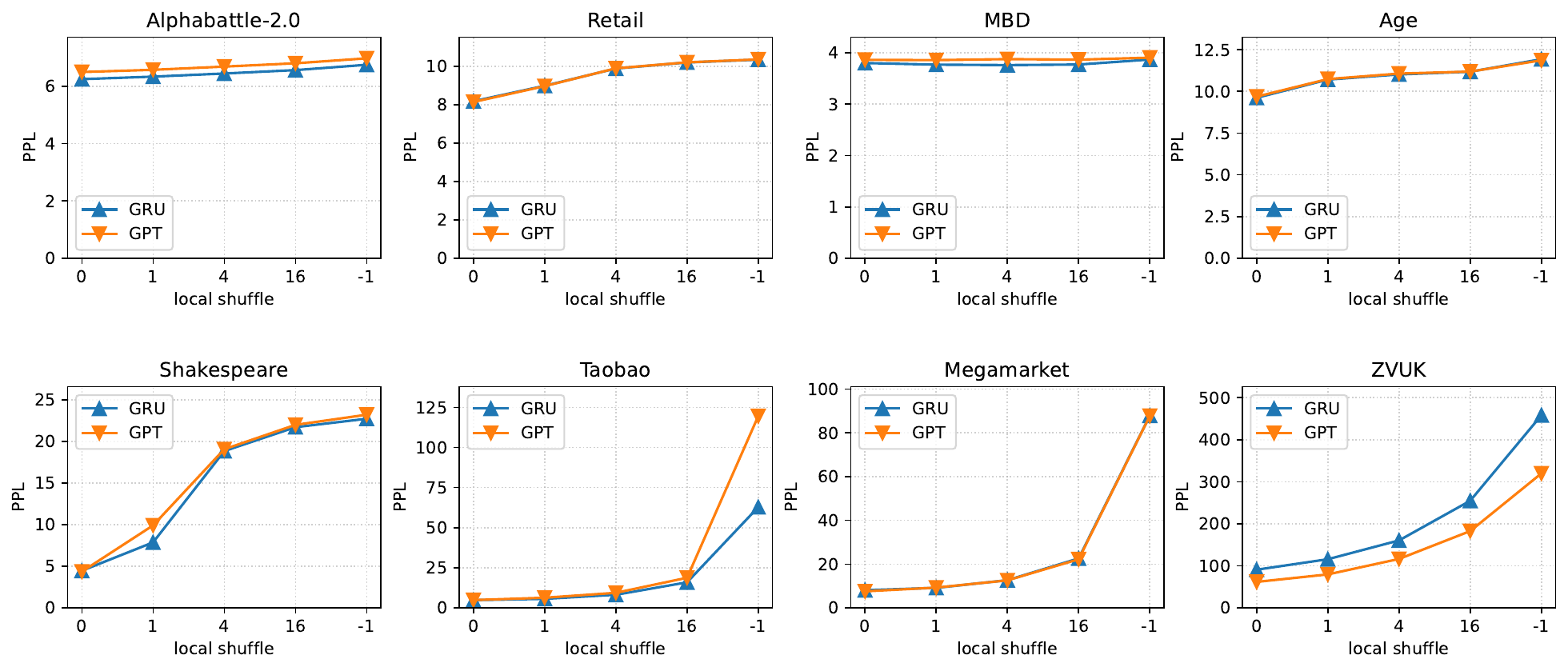}
\end{center}
\caption{Next $N$ tokens forecasting. Perplexity results.}
\label{fig:perplexity_score_on_all_datasets}
\end{figure}

\begin{figure}[h]
\begin{center}
\includegraphics[width=1\linewidth]{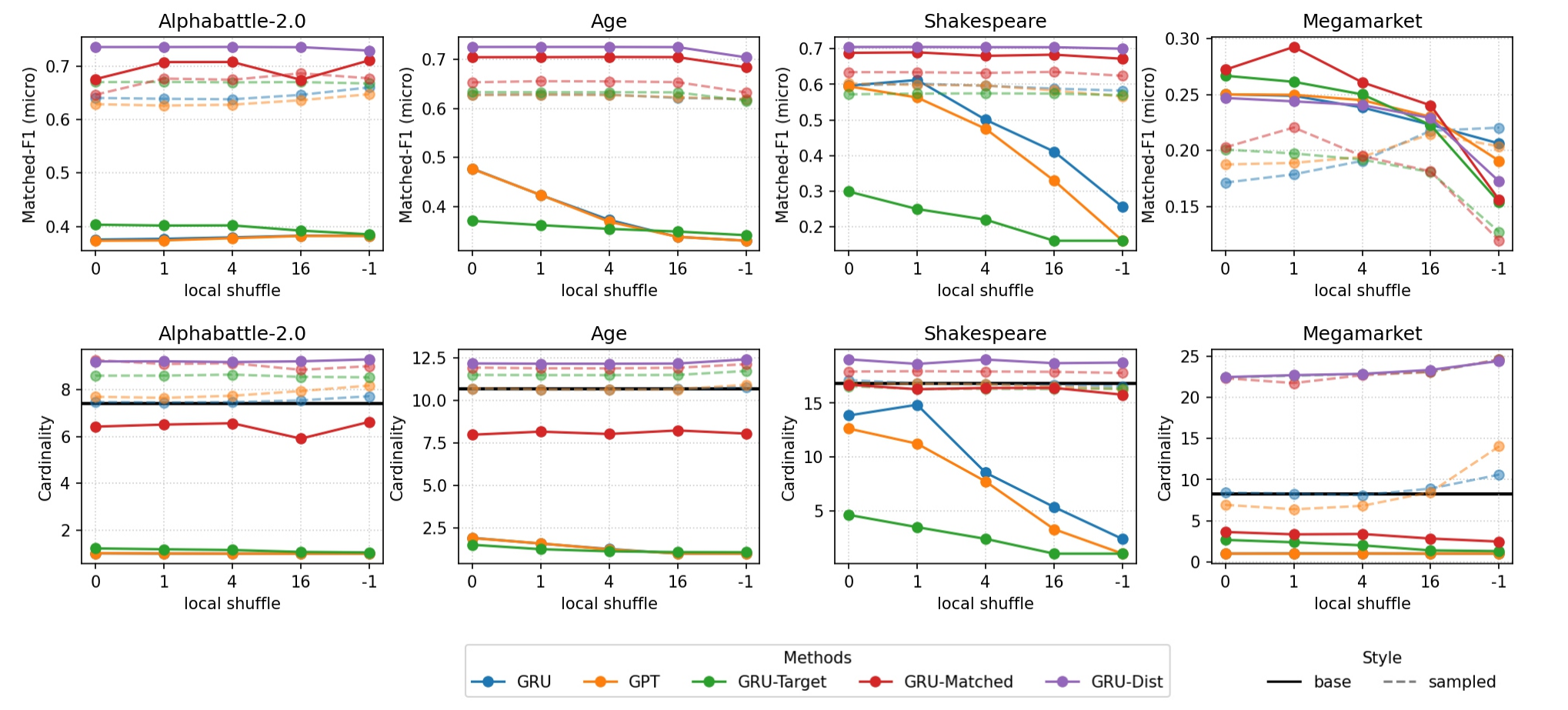}
\end{center}
\caption{Effect of Local Event Shuffling on Model Performance. We report Matched-F1 score and Carnality for four datasets. Results for other datasets and metrics can be found in Appendix~\ref{appendix:all_results}}
\label{fig:f1_and_crd_on_all_datasets}
\end{figure}

\textbf{Local permutation experiments further corroborate} these findings; results are shown in Figure~\ref{fig:f1_and_crd_on_all_datasets}. Shakespeare and Zvuk exhibit sharp performance degradation when sequences are shuffled, indicating strong local sequential structure. In contrast, most banking datasets show little to no degradation, reflecting the irrelevance of event order. This trend is especially evident in Figure~\ref{fig:perplexity_score_on_all_datasets}, which illustrates minimal perplexity degradation under shuffling for these datasets.

\textbf{Model performance.}
The order-invariant model \textit{GRU-Dist} achieves the best overall performance on most datasets, with the notable exceptions of Taobao and Megamarket. Further analysis reveals that these two datasets contain long-horizon repetitive patterns of identical events. This characteristic aligns with the observation that the \textbf{repeat} baseline performs best on them, as it effectively exploits such redundancy.

\section{Conclusion}

Our study demonstrates that model performance in event-sequence forecasting is tightly linked to dataset characteristics.  
\textit{GRU-Target} performs best when temporal order is largely irrelevant, while \textit{GRU-Matched} and \textit{GRU-Dist} consistently outperform other approaches; in particular, \textit{GRU-Dist} is considered as more appropriate baseline for banking tasks, where the presence of events is more informative than their order. The drop in GRU-Dist performance on Megamarket and Taobao is likely due to dataset-specific characteristics: sequences often contain long run of identical tokens, which is difficult to reproduce when sampling from a categorical distribution. GPT-based models, by contrast, are more effective on datasets sensitive to local permutations, such as text. Therefore, in scenarios where temporal order is essential, Next-Token Prediction (NTP) and Multi-Token Prediction remain the preferable approaches.

Cardinality also proves to be a useful diagnostic of mode collapse: in datasets like Shakespeare, shuffling removes structural cues and autoregressive models degenerate to the modal category. More broadly, when no meaningful local ordering exists, models tend to collapse to the mode (Figure~\ref{fig:f1_and_crd_on_all_datasets}).

Taken together, these results highlight the value of simple dataset-level diagnostics (exponential decay parameter $\lambda$, staticity, cardinality) for anticipating model behavior, and demonstrate the advantages of order-invariant objectives in domains such as retail and banking, where event presence matters more than sequence order.

Indeed, it is worth noting that the proposed \textit{GRU-Dist} method can be extended from single-category forecasting to multi-feature prediction through cascade modeling.

\textbf{Acknowledgment on LLM assisted writing:} This paper used open access Qwen3-Max, in some parts of the paper, for proofreading and text rephrasing in accordance with formal style.



\bibliography{iclr2026_conference}
\bibliographystyle{iclr2026_conference}

\appendix
\section{Appendix}

\subsection{HPO details}
\label{app:hpo}
For hyperparameter optimization (HPO), we use Optuna~\citep{akiba2019optuna} with the Tree-structured Parzen Estimator (TPE) sampler.  
For each model–dataset pair, we allocate an HPO budget of 100 training runs, capping the total computational cost at 18 NVIDIA A100 GPU-days.  
We reserve 15\% of the training set as a validation subset for early stopping and hyperparameter selection.  
The best-performing hyperparameters are then used to train the final model for evaluation and all subsequent study experiments.

\subsection{Local Global temporal Invariance}
\begin{figure}[ht]
    \centering
    \includegraphics[width=0.9\linewidth]{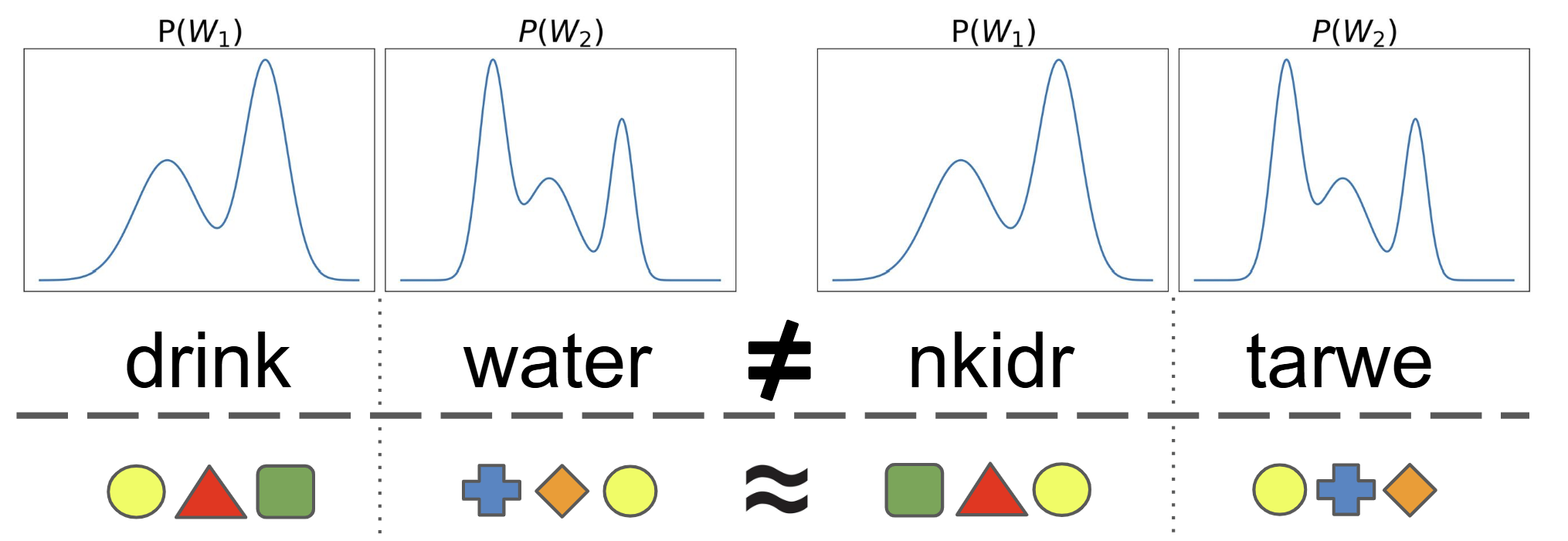}
    \caption{Example how order importance differs in different types of data. Even though in both cases horizon distribution doesnt change, event sequence still make sence after permut inside intervals.}
    \label{fig:order_example}
\end{figure}

In Figure~\ref{fig:order_example} we illustrate local / global invariance.

\subsection{Datasets description and preprocessing}
\label{app:datasets}

\paragraph{MBD}\footnote{\url{https://huggingface.co/datasets/ai-lab/MBD}} is a multimodal banking dataset introduced in~\cite{mollaev2024multimodal}. The dataset contains an industrial-scale number of sequences, with data from more than 1.5 million clients in 2 year period. Each client corresponds to a sequence of events. This multi-modal dataset includes card transactions, geo-position events, and embeddings of dialogs with technical support. For our analysis, we use only card transactions. We use a temporal train–test split: transactions from the first year form the training set, and those from the second year form the test set.

\paragraph{Age} dataset\footnote{\url{https://ods.ai/competitions/sberbank-sirius-lesson}} consists of 44M anonymized credit card transactions representing 50K individuals.
The target is to predict the age group of a cardholder that made the transactions.
Each transaction includes the date, type, and amount being charged.
The dataset was first introduced in scientific literature in work~\cite{babaev2022coles}. We perform a user-based split: 80\% of sequences are assigned to the training set, and the remaining 20\% of sequences are held out for testing.

\paragraph{Retail} dataset\footnote{\url{https://ods.ai/competitions/x5-retailhero-uplift-modeling}} comprises 45.8M retail purchases from 400K clients, with the aim of predicting a client's age group based on their purchase history.
Each purchase record includes details such as time, item category, the cose, and loyalty program points received.
The age group information is available for all clients, and the distribution of these groups is balanced across the dataset.
The dataset was first introduced in scientific literature in work~\cite{babaev2022coles}. We perform a user-based split: 80\% of sequences are assigned to the training set, and the remaining 20\% of sequences are held out for testing.

\paragraph{Alphabattle-2.0} datase~\footnote{\url{https://www.kaggle.com/datasets/mrmorj/alfabattle-20}} The AlfaBattle 2.0 dataset contains bank customers' transaction records over two years, with the goal of predicting loan default based on behavioral history. Each record includes 18 features (3 numeric, 15 categorical) per transaction. We use the official test split provided by the dataset creators.

\paragraph{Shakespeare}
Dataset consists of character-level text extracted from Shakespeare’s works, preprocessed into individual speech segments. Each speech is tokenized using a vocabulary of unique characters mapped to integer codes based on frequency. The final dataset is split into train and test sets (80/20). The dataset is designed for character-level language modeling and was selected due to it obvious temporal importance.

\paragraph{Zvuk} dataset\footnote{\url{https://www.kaggle.com/datasets/alexxl/zvuk-dataset}} is introduced in~\citeyear{shevchenko2024variability} and contains 244.7M music listening events grouped into 12.6M sessions from 382K users, recorded during the same five-month period (January–May 2023).
In total, it spans 1.5M unique tracks.
Each record includes a user ID, session ID, track ID, timestamp, and play duration (considering only plays covering at least 30\% of track length).
The dataset is tailored to music consumption, excluding podcasts and audiobooks, and enables evaluation of recommendation models in domains with stronger sequential dynamics. We use a temporal train–test split: transactions from the first two months form the training set, and other two month form the test set.

\paragraph{MegaMarket} dataset\footnote{\url{https://www.kaggle.com/datasets/alexxl/megamarket?select=megamarket.parquet}} is introduced in~\citeyear{shevchenko2024variability} and comprises 196.6M user interactions collected over a five-month period (January–May 2023).
It covers 2.7M users, 3.56M items, and 10,001 product categories, with events including views, favorites, cart additions, and purchases.
Each record contains a user ID, item ID, event type, category ID, timestamp, and normalized price.
The dataset represents large-scale e-commerce behavior and is intended for sequential recommendation tasks. This dataset follows the same temporal train/test split as Zvuk.

\paragraph{Taobao}\footnote{\url{https://tianchi.aliyun.com/dataset/46}} The dataset comprises user behaviors from Taobao, including clicks, purchases, adding items to the shopping cart, and favoriting items. These events were collected between November 18 and December 15. The training set encompasses data from November 18 to December 1, while the test set includes clicks from December 2 to December 15.

\subsection{Staticity index plots for key datasets}
\label{appendix:staticity}

\begin{figure}[ht]
  \centering
    \includegraphics[width=\linewidth]{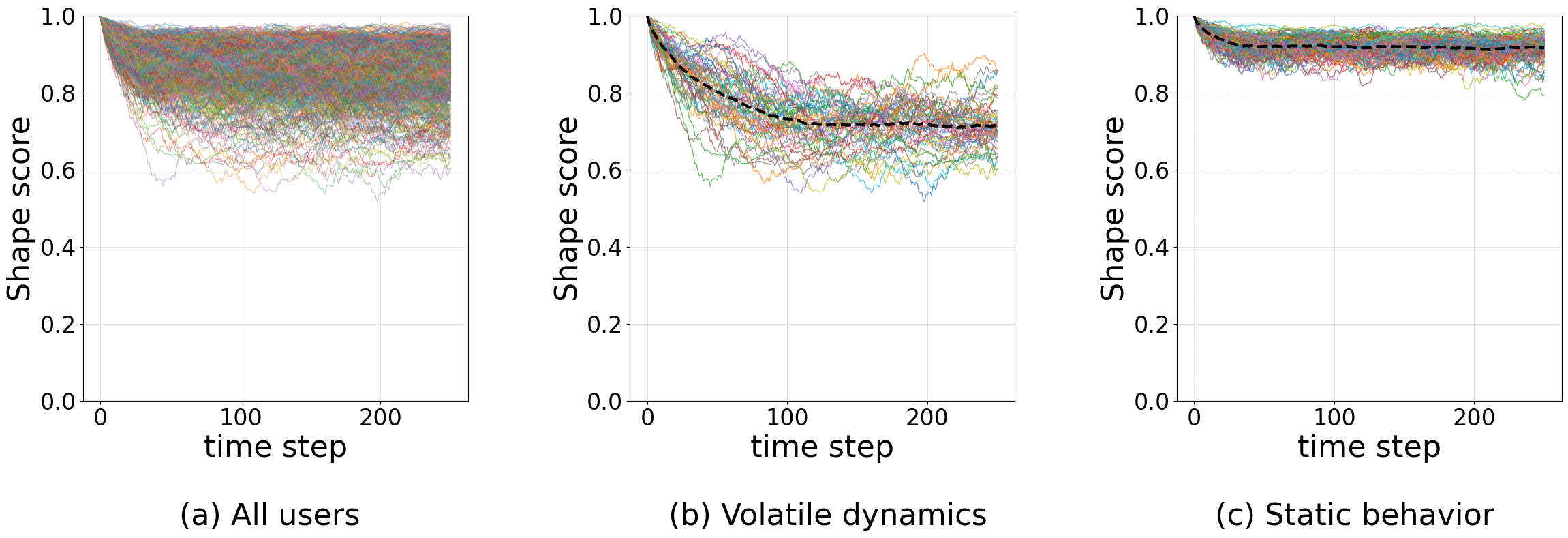}
  \caption{Shape score drift for MBD dataset}
  \label{fig:staticity-mbd}
\end{figure}

\begin{figure}[ht]
  \centering
    \includegraphics[width=\linewidth]{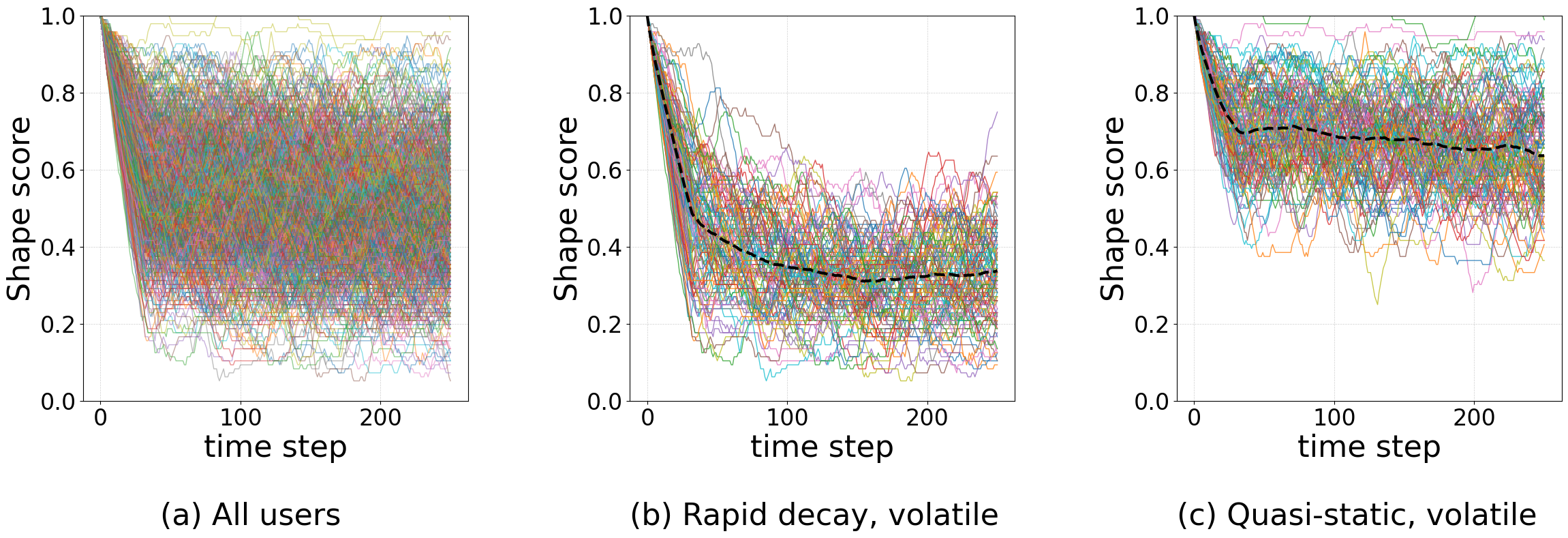}
  \caption{Shape score drift for ZVUK dataset}
  \label{fig:staticity-zvuk}
\end{figure}

\begin{figure}[ht]
  \centering
    \includegraphics[width=\linewidth]{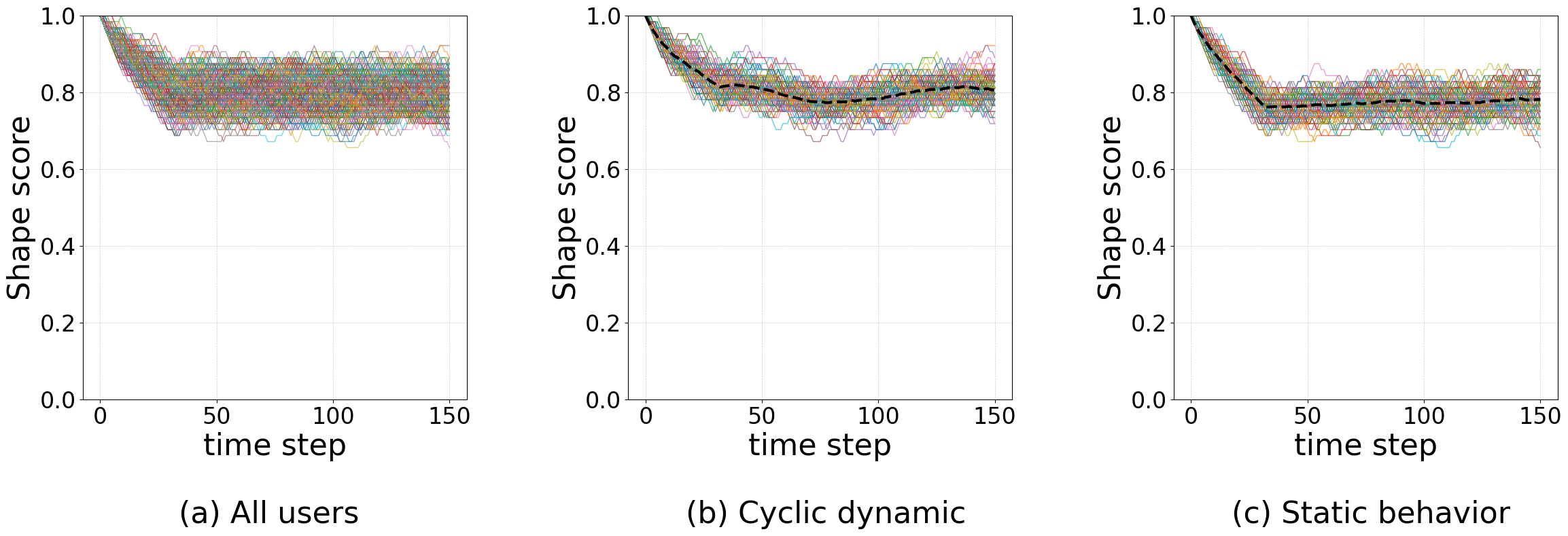}
  \caption{Shape score drift for Shakespeare dataset}
  \label{fig:staticity-shakespeare}
\end{figure}

For each dataset, we compute drift trajectories for all sequence and cluster them into a small number of groups with internally consistent dynamics (Figure~\ref{fig:staticity-mbd}–\ref{fig:staticity-shakespeare}). Across banking datasets (MBD, Retail, Age, Alphabattle) the dominant clusters are static, as exemplified for \textbf{MBD} (Figure~\ref{fig:staticity-mbd}c), these clusters exhibit negligible temporal drift. For such sequences, learning the user’s category distribution suffices to forecast the next block of events. Trajectories with pronounced drift are rare. In MBD specifically, such sequences are observed in fewer than 6\% of users (Figure~\ref{fig:staticity-mbd}b).

In contrast to banking datasets, recommender–system data exhibit much greater variability. In \textbf{ZVUK} (Figure~\ref{fig:staticity-zvuk}), two characteristic regimes dominate: one cluster shows a sharp initial drop from the baseline followed by persistent high-variance fluctuations, while another appears quasi-static yet remains noisy around its trend. Such patterns reflect the broader nature of recommender logs: users interact with a large and diverse sets of items, and their behavior shifts more frequently than in retail domains where event types are limited and highly regular. And as a consequence, their later-window distributions are more clearly separated from the first-window distribution.

The outlier in this collection is the \textbf{Shakespeare} text dataset (Figure~\ref{fig:staticity-shakespeare}). Although it is non-transactional, its dynamics resemble banking data more than recommender logs: drift trajectories are mostly flat and volatility remains low. At the same time, weak periodic or gradual shifts are observable, indicating that the sequences are not fully static but display a modest degree of temporal variation.

\subsection{Features Impact in Category Forecasting Quality}

We investigated whether predicting a target feature benefits more from incorporating the full feature vector or from relying exclusively on its own historical values.

On the MBD dataset, experiments in the \emph{All‐to‐One} and \emph{One‐to‐One} modes reveal that the autoregressive model’s performance degrades when exposed to complete with the complete feature vector. The additional inputs act as noise, impeding the model’s ability to reproduce the mode of the target distribution. In the \emph{One‐to‐One} mode—where the model sees only the history of the target feature—it easily learns the mode and reports a formal increase in accuracy; however, this gain is illusory, as the generated sequences become overly uniform and lack realism~\ref{tab:all-to-one}.

\begin{table}[ht]
\centering
\caption{Effect of training with all tokens vs. event type only (\textit{Matched\mbox{-}F1 micro}).}
\label{tab:all-to-one}
\begin{tabular}{l c}
\hline
\textbf{Dataset} & \textbf{Change (\%)} \\
\hline
MBD        & $+2.85$ \\
AGE        & $-24.94$ \\
MM         & $+13.66$ \\
\hline
\end{tabular}
\end{table}

By contrast, on datasets with a strong sequential structure, such as \emph{Megamarket}, the opposite pattern emerges. The autoregressive mechanism leverages ordering information and, when augmented with additional features, predicts beyond mere modal values, resulting in a significant improvement in performance metrics.

\subsection{Metrics}

\subsubsection{Matched\mbox{-}F1 micro}
\label{appendix:f1}

\paragraph{Precision and recall.}
\[
  \text{Prec}_k = 
    \frac{\text{TP}_k}{\text{TP}_k+\text{FP}_k},
  \qquad
  \text{Rec}_k =
    \frac{\text{TP}_k}{\text{TP}_k+\text{FN}_k}.
\]

\paragraph{Macro averaging.}
\[
  F1_{\text{macro}}
  = \frac{1}{K}\sum_{k=1}^{K}
      \frac{2\,\text{Prec}_k\,\text{Rec}_k}
           {\text{Prec}_k+\text{Rec}_k}.
\]
Each class contributes equally; the score is sensitive to rare categories.

\paragraph{Micro averaging.}
Aggregating counts over classes,

\begin{align}
\text{TP} &= \sum_k \text{TP}_k, &
\text{FP} &= \sum_k \text{FP}_k, &
\text{FN} &= \sum_k \text{FN}_k,
\end{align}

\begin{align}
F1_{\text{micro}} &= \frac{2\,\text{TP}}{2\,\text{TP} + \text{FP} + \text{FN}}.
\end{align}

This variant weights categories by frequency and reflects overall
throughput.

\subsubsection{Cardinality metric.}
\label{appendix:cardinality}
Let \(G_i = \bigl(x_{t+1}^{(i)},\dots,x_{t+L}^{(i)}\bigr)\) denote the
\(L\)-step segment generated for sequence \(i\) and
\(\mathcal{C}(G_i)=\bigl\{x\in G_i\bigr\}\) the set of
\emph{distinct} categories appearing in that segment.
We define the per‑sequence cardinality as
\[
  C_i \;=\; \bigl|\mathcal{C}(G_i)\bigr|.
\]
The dataset‑level score is the average
\[
  \mathrm{Cardinality}
  \;=\;
  \frac{1}{N}\sum_{i=1}^{N} C_i,
\]
where \(N\) is the number of sequences under evaluation.
An \emph{overall} variant first concatenates all generated segments,
\(\tilde G=\bigcup_i G_i\), and reports
\(C_\mathrm{overall}=|\mathcal{C}(\tilde G)|\).

\textbf{Purpose.}
Cardinality captures the \emph{category diversity} produced by a model:
low values signal mode collapse, whereas values close to the ground‑truth
cardinality indicate faithful reproduction of event variety.
We compute the metric for both generated (\(C_\text{gen}\)) and reference
(\(C_\text{orig}\)) sequences, allowing direct comparison of a model’s
diversity against empirical data.

\subsection{Neural Backbone Architectures}

We evaluate two neural backbone architectures for sequence modeling:

\begin{itemize}
  \item \textbf{GRU:} A gated recurrent unit (GRU) network excels at capturing local dependencies and stationary patterns in short to moderately long time series.
  \item \textbf{GPT:} A self-attention–based model capable of modeling long-range dependencies, crucial for sequences with complex contextual interactions and implicit event relationships.
\end{itemize}

\subsection{Additional Results}
\label{appendix:all_results}
\begin{figure}[h]
\begin{center}
\includegraphics[width=\linewidth]{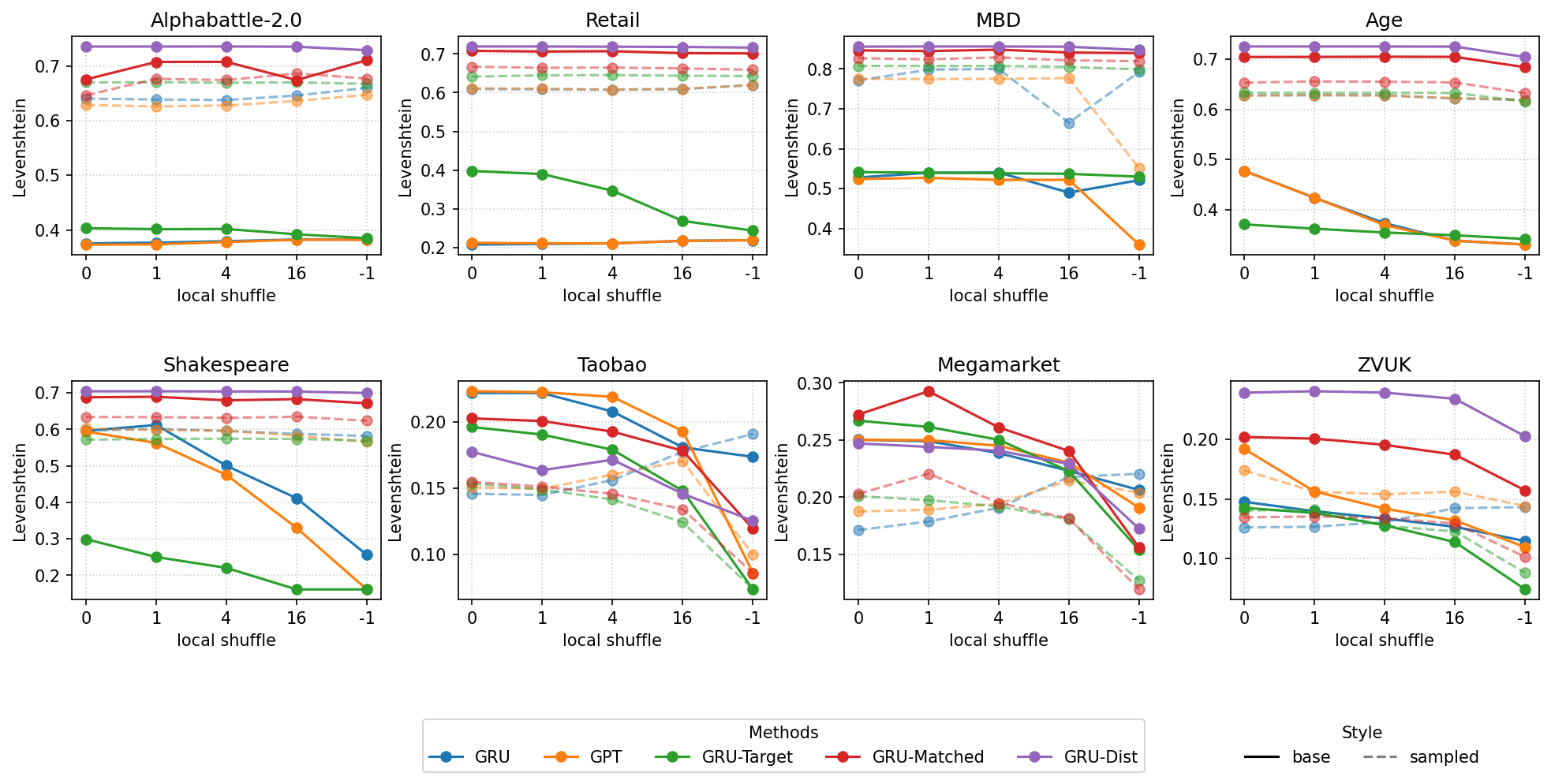}
\end{center}
\caption{Levenshtein score on all datasets.}
\label{fig:lev_score_on_all_datasets}
\end{figure}

\begin{figure}[h]
\begin{center}
\includegraphics[width=1\linewidth]{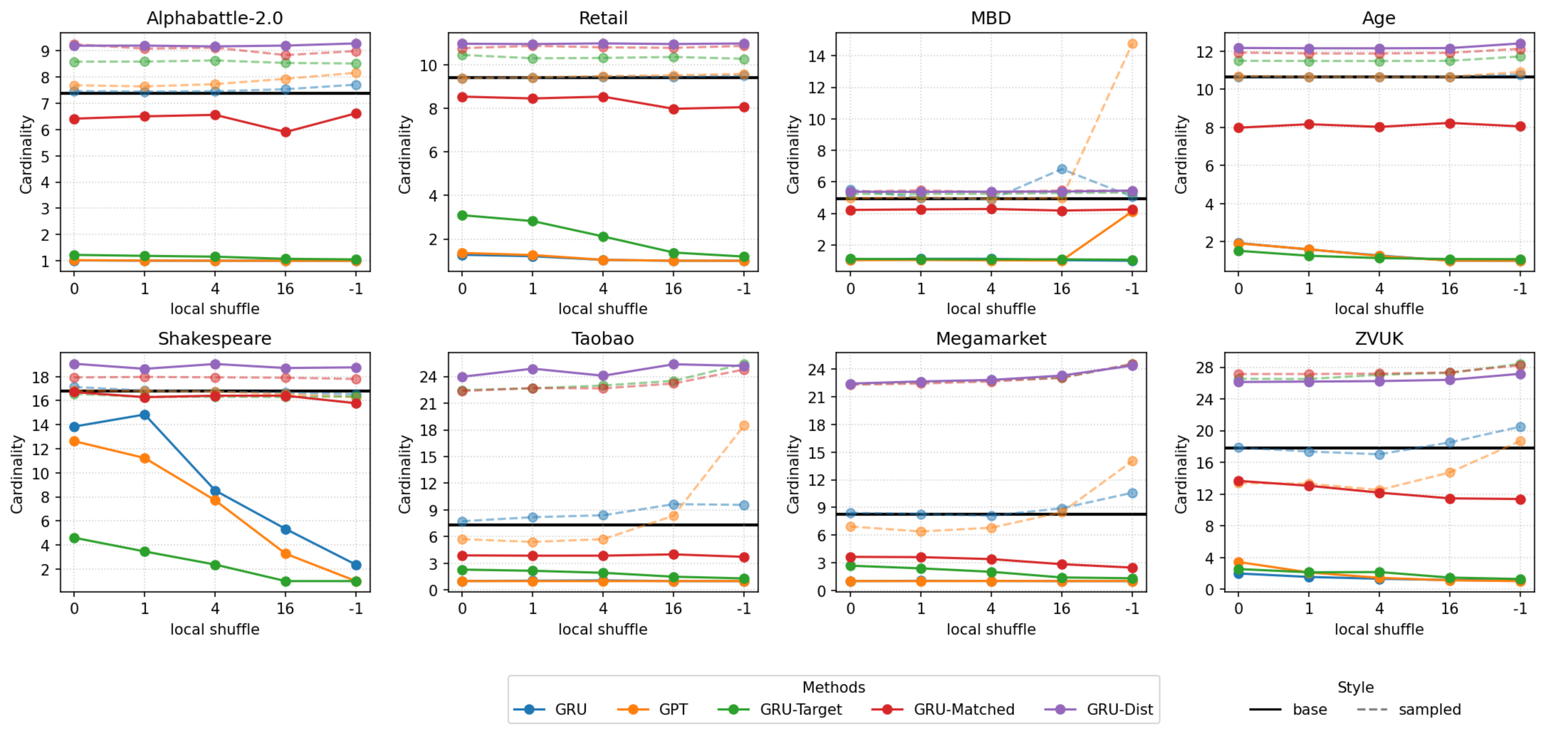}
\end{center}
\caption{Effect of local shuffle on cardinality.}
\label{fig:cardinality_on_all_datasets}
\end{figure}

\begin{figure}[h]
\begin{center}
\includegraphics[width=\linewidth]{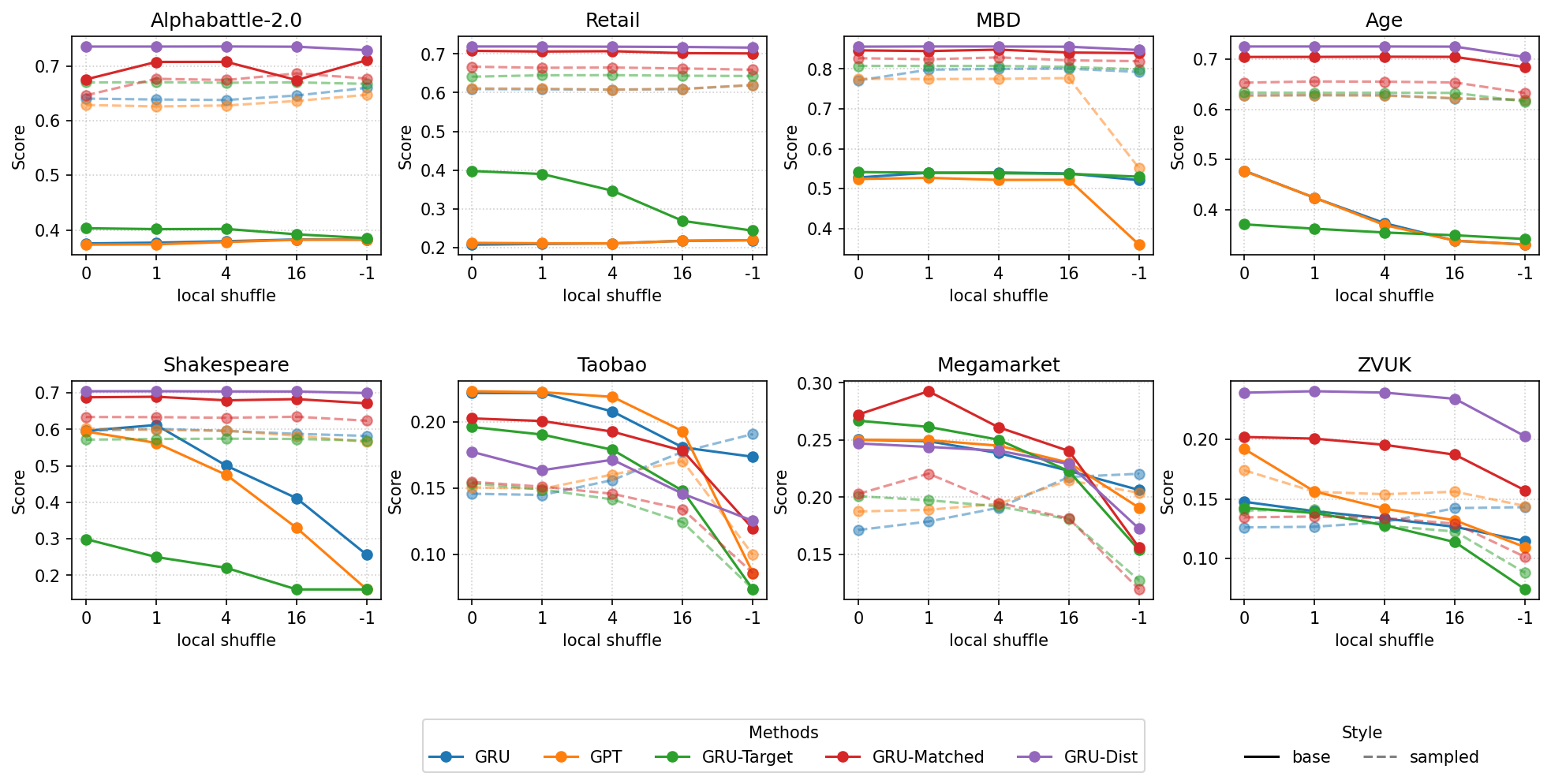}
\end{center}
\caption{Next $N$ tokens forecasting. \textit{Matched\mbox{-}F1 (micro)} results.}
\label{fig:f1_score_on_all_datasets}
\end{figure}

For completeness, we report all evaluation metrics across datasets.  
Levenshtein distance is included as an order-sensitive measure to quantify degradation under local shuffling (Figure~\ref{fig:lev_score_on_all_datasets}), while the effect of shuffling on category diversity is illustrated by cardinality (Figure~\ref{fig:cardinality_on_all_datasets}).  
The main text focuses on the order-invariant \textit{Matched-F1 (micro)} (Figure~\ref{fig:f1_score_on_all_datasets}), which we adopt as the primary evaluation metric throughout the study.

\end{document}